\documentclass[10pt,twocolumn,letterpaper]{article}

\usepackage[pagenumbers]{wacv} 

\usepackage{graphicx}
\usepackage{amsmath}
\usepackage{amssymb}
\usepackage{booktabs}
\usepackage{import}
\usepackage{makecell}
\usepackage{multirow}
\usepackage{url}
\usepackage{threeparttable}
\usepackage[dvipsnames]{xcolor}
\usepackage{caption}
\usepackage{wrapfig}
\usepackage{xspace}
\usepackage{pifont}
\newcommand{\cmark}{\ding{51}}%
\newcommand{\xmark}{\ding{55}}%
\usepackage{tcolorbox}
\tcbset{colback=white}
\colorlet{titleblue}{blue!80!black}
\colorlet{titlered}{red!80!black}
\colorlet{titlegreen}{green!80!black}
\usepackage{soul}

\usepackage[pagebackref,breaklinks,colorlinks]{hyperref}

\usepackage[capitalize]{cleveref}
\crefname{section}{Sec.}{Secs.}
\Crefname{section}{Section}{Sections}
\Crefname{table}{Table}{Tables}
\crefname{table}{Tab.}{Tabs.}

\def\wacvPaperID{*****} 
\def\confName{WACV}
\def\confYear{2025}

\begin{document}


\author{%
  Yunsheng Ma\thanks{Work conducted as an intern at Toyota InfoTech Labs.} $^{1,2}$, Amr Abdelraouf$^1$, Rohit Gupta$^1$, Ziran Wang$^2$, Kyungtae Han$^1$\\
  $^{1}$ Toyota InfoTech Labs, Mountain View, CA, USA\\
  $^{2}$ Purdue University, West Lafayette, IN, USA\\
  \texttt{\small \{yunsheng,ziran\}@purdue.edu} \texttt{\small \{amr.abdelraouf,rohit.gupta,kt.han\}@toyota.com} 
}

\title{Video Token Sparsification for Efficient Multimodal LLMs in Autonomous Driving}

\maketitle

\begin{abstract}
Multimodal large language models (MLLMs) have demonstrated remarkable potential for enhancing scene understanding in autonomous driving systems through powerful logical reasoning capabilities. However, the deployment of these models faces significant challenges due to their substantial parameter sizes and computational demands, which often exceed the constraints of onboard computation. One major limitation arises from the large number of visual tokens required to capture fine-grained and long-context visual information, leading to increased latency and memory consumption. To address this issue, we propose Video Token Sparsification (VTS), a novel approach that leverages the inherent redundancy in consecutive video frames to significantly reduce the total number of visual tokens while preserving the most salient information. VTS employs a lightweight CNN-based proposal model to adaptively identify key frames and prune less informative tokens, effectively mitigating hallucinations and increasing inference throughput without compromising performance.
We conduct comprehensive experiments on the DRAMA and LingoQA benchmarks, demonstrating the effectiveness of VTS in achieving up to a 33\% improvement in inference throughput and a 28\% reduction in memory usage compared to the baseline without compromising performance.
\end{abstract}


\section{Introduction}
\label{sec:intro}
Autonomous driving has undergone significant advancements in recent years, transitioning from modular pipelines to end-to-end driving models~\cite{chen_end--end_2024}. Despite these rapid developments, existing end-to-end frameworks still face two major limitations. First, they lack the ability to handle corner cases effectively in a human-like manner. Second, they lack natural language capabilities, specifically the ability to explain or justify their actions to human users and follow human instructions~\cite{echterhoff2024driving}. These limitations constrain applicability in real-world deployments and hinder user trust and user-friendliness.

The emergence of multimodal large language models (MLLMs)~\cite{liu_visual_2023} has catalyzed a revolution in autonomous driving, driven by their advanced cognitive and logical reasoning capabilities. Recent studies have explored leveraging MLLMs to develop autonomous driving systems that enable language-based scene descriptions (e.g., enabling chain-of-thought reasoning) and language-conditioned planning following human instructions. For example, DriveLM~\cite{sima_drivelm_2024} considers graph visual question answering, where question-answer pairs are interconnected via logical dependencies at the object level and task level. LMDrive~\cite{shao_lmdrive_2024} proposes an end-to-end, closed-loop, language-based autonomous driving framework that interacts with the dynamic environment via multimodal sensor data and natural language instructions. These approaches aim to address the limitations of existing end-to-end frameworks by incorporating natural language understanding and generation capabilities.

However, the introduction of MLLMs in autonomous driving systems also presents two significant challenges. First, the substantial parameter sizes and computational demands of MLLMs often exceed the constraints of onboard computation in autonomous vehicles, hindering their practical deployment and real-time performance. While model distillation can be employed to fit the onboard constraints, it can inevitably hinder the reasoning capabilities of MLLMs~\cite{wei_chain--thought_2022}. Second, capturing fine-grained and long-context visual information requires a large number of visual tokens, further exacerbating the computational burden. Existing works mainly focus on single-image token reduction~\cite{bolya_token_2023}. Such strategies naturally cannot leverage the redundant information in consecutive video frames and may lose critical visual information when applying a high compression rate.

To address these challenges and enable the practical deployment of MLLMs in autonomous driving systems, we propose Video Token Sparsification (VTS), a novel approach that leverages the redundancy in successive video frames to reduce the excessive visual tokens while preserving the most salient information. The key contributions of our work are as follows:
\begin{itemize}
\item We propose Video Token Sparsification (VTS), a novel approach that adaptively reduces the number of visual tokens by exploiting the redundancy in consecutive video frames and identifying the most informative tokens.
\item We introduce a lightweight proposal model for efficient key frame selection and inter-frame token pruning, enabling VTS to operate with minimal computational overhead.
\item We conduct comprehensive experiments on the DRAMA and LingoQA benchmarks, demonstrating the effectiveness of VTS in removing 40\% of redundant visual tokens and improving inference throughput by up to 33\% without compromising performance on various video question answering tasks in autonomous driving scenarios.
\end{itemize}



\section{Related Works}
\label{sec:related}
\subsection{Vision-Language Models for Driving}
Multimodal Large Language Models (MLLMs) have gained attention in autonomous driving due to their advanced reasoning capabilities. Studies have explored leveraging MLLMs for developing autonomous driving systems~\cite{sima_drivelm_2024,mao_language_2023,xu_drivegpt4_2023,cui_survey_2024}, but many open questions remain.
Research on VLM architectures for autonomous driving includes OmniDrive~\cite{wang_omnidrive_2024}, which proposes a 3D MLLM architecture using sparse queries for perception-action alignment, and DriveVLM~\cite{tian_drivevlm_2024}, which integrates chain-of-thought modules for scene understanding and planning. Reconciling Bird's-Eye-View (BEV) scene representations with language-based descriptions is another important direction. CLIP-BEVFormer enhances BEV backbones using contrastive learning, while TOKEN~\cite{tian_tokenize_2024} tokenizes the world into object-level knowledge for improved reasoning. Language-conditioned planners, such as LMDrive~\cite{shao_lmdrive_2024} and LaMPilot~\cite{ma_lampilot_2024}, integrate natural language instructions with multi-modal sensor data and generate code leveraging functional primitives, respectively. Despite the potential of MLLMs in autonomous driving, their computational complexity and memory requirements pose challenges for real-world deployment. The proposed Video Token Sparsification (VTS) approach addresses these challenges by efficiently reducing visual tokens while maintaining high performance.

\subsection{Token Reduction}
Token reduction techniques can be categorized into token merging and token pruning, and are proven effective for improving computation efficiency. Token merging combines similar or redundant tokens, while token pruning removes less informative or irrelevant tokens. For token merging, ToMe~\cite{bolya_token_2023} merges similar parts in each block of Vision Transformers (ViTs). ToMeSD~\cite{bolya_token_2023-1} and VidToMe~\cite{li_vidtome_2024} extend this idea to speed up diffusion models and enhance temporal consistency in generated videos, respectively.
In the token pruning category, MADTP~\cite{cao_madtp_2024} introduces a multimodal dynamic token pruning strategy for Vision-Language Models. The Hourglass Tokenizer (HoT)~\cite{li_hourglass_2024} presents a pruning-and-recovering framework for efficient transformer-based 3D human pose estimation. PaPr~\cite{mahmud_papr_2024} proposes a method for pruning redundant patches using lightweight convolutional neural networks. The Video Token Sparsification (VTS) approach proposed in this paper falls into the token pruning category and is the first to apply token reduction techniques for multimodal LLMs in the autonomous driving domain. VTS leverages the redundancy in consecutive video frames to adaptively prune less informative tokens while preserving the most salient driving information.

\section{Methodology}
\begin{figure*}[!t]
    \centering
    \includegraphics[width=0.65\linewidth]{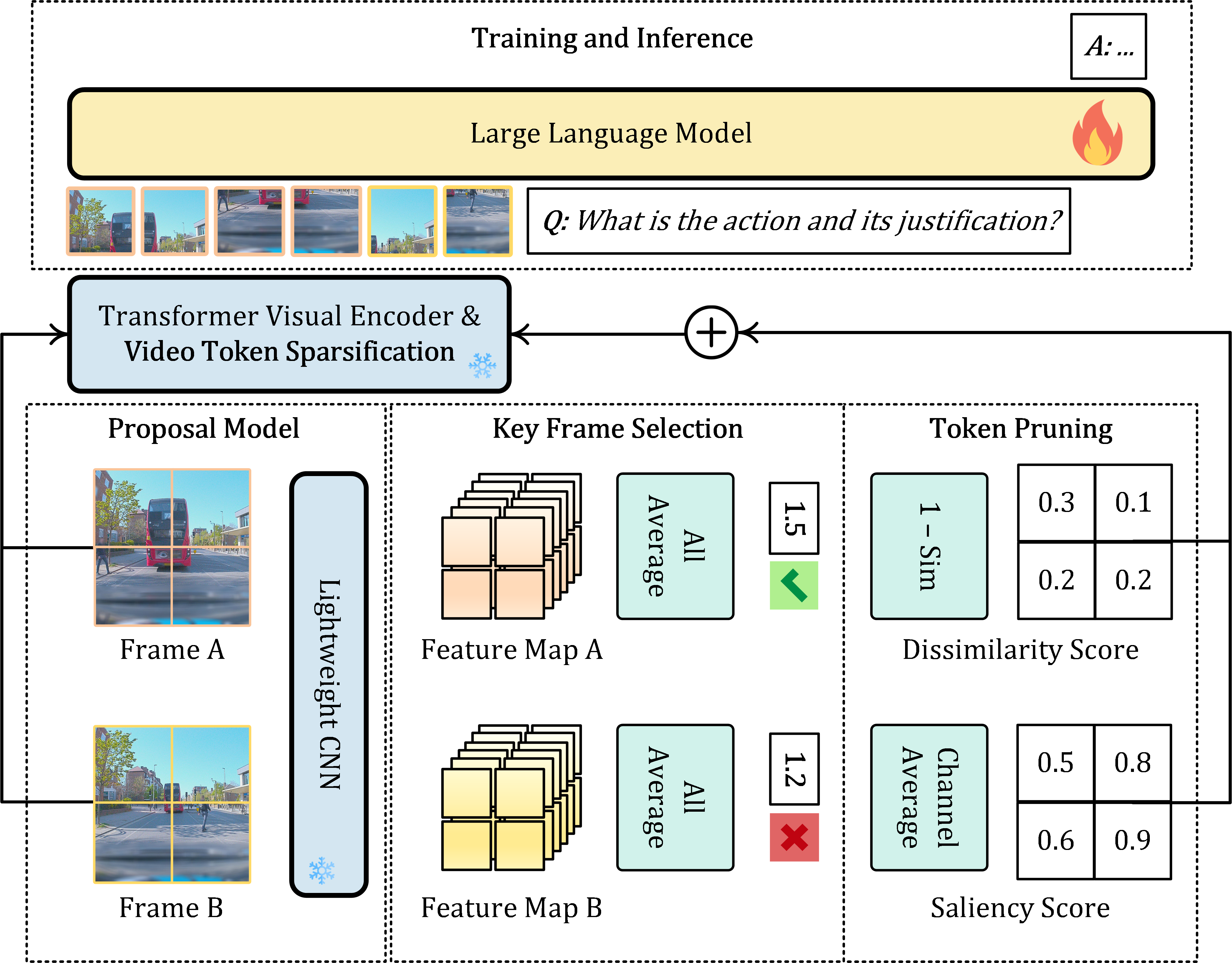}
    \caption{\textbf{Overview of Video Token Sparsification (VTS)}: The input consists of both the current frame and previous frames (referred to as Frame A and Frame B in this example). A CNN-based proposal model generates feature maps for each frame. VTS identifies a key frame (Frame A) based on these feature maps and computes a pruning score for each token in the non-key frames (Frame B), considering both saliency and dissimilarity to the corresponding tokens in the key frame. The top $s\%$ of tokens from the non-key frames are selected and concatenated with the tokens from the key frame to form the final sparsified token sequence, which is then input to the LLM for efficient reasoning.}
    \label{fig:method}
    \vspace{-3mm}
\end{figure*}

\label{sec:method}
\subsection{Temporal Reasoning}
Temporal reasoning is crucial for autonomous driving systems to safely navigate dynamic environments. Human drivers inherently consider the historical context when interpreting the current scene and deciding on appropriate actions. For example, observing the trajectory of an approaching vehicle allows inferring its likely future path and intentions. Incorporating such temporal reasoning capabilities is essential for autonomous driving systems.

Our approach enhances the reasoning capabilities of the MLLM by providing a temporal context in the form of multiple consecutive input frames. As shown in \cref{fig:method}, for a single MLLM request at time $t$, the input consists of the current frame $\mathbf{X}_t\in\mathbb{R}^{H\times W\times 3}$ and the $M$ previous frames $\mathbf{X}_{t-M:t-1}\in\mathbb{R}^{M\times H\times W\times 3}$.
To extract visual features, we employ a pretrained visual encoder $\mathfrak{g}$, such as a ViT~\cite{dosovitskiy_image_2021}. The encoder maps each input frame $\mathbf{X}_i$ to a sequence of visual tokens $\mathbf{H}_i = \mathfrak{g}(\mathbf{X}_i)\in \mathbb{R}^{N\times D}$, where $N=H\times W/p^2$ is the number of tokens determined by the patch size $p$, and $D$ is the embedding dimension.

This multi-frame input allows the MLLM to reason about the temporal evolution of the scene and make more informed decisions. 
However, naively increasing the temporal context by adding more frames leads to a proportional increase in the number of input tokens and computational cost. In the following sections, we introduce our video token sparsification approach to effectively leverage the temporal redundancy and reduce the number of tokens while preserving salient information.

\subsection{Video Token Sparsification (VTS)}
\subsubsection{Proposal Model}
The central idea of our approach is to identify and retain only the most informative visual tokens across the frames in the input. We draw inspiration from the observation that lightweight convolutional neural networks (CNNs) are effective at efficiently identifying salient image regions~\cite{mahmud_papr_2024}. While these networks may lack the capacity for fine-grained classification, their convolutional layers can serve as a computationally efficient feature extractor to guide the token selection process.

We introduce a lightweight CNN $\mathfrak{p}$ as the proposal model which generates a feature map $\mathbf{F}_t\in\mathbb{R}^{H/q\times W/q\times C}$ for each input frame $\mathbf{X}_t$:
\begin{equation}
\mathbf{F}_{t-M:t}=\mathfrak{p}(\mathbf{X}_{t-M:t}),
\end{equation}
where $q$ is the spatial downsampling factor determined by the CNN architecture, and $C$ is the number of output channels.

To align the spatial dimensions of the feature map $\mathbf{F}_t$ with those of the visual tokens $\mathbf{H}_t$, we apply a simple bicubic interpolation operation $U:\mathbb{R}^{H/q\times W/q}\mapsto\mathbb{R}^{H/p\times W/p}$. This results in an upsampled feature map $\mathbf{P}_t=U(\mathbf{F}_t)$ that associates each visual token with a corresponding $C$-dimensional feature vector.

\subsubsection{Key Frame Selection}
Given the feature vectors computed by the proposal model, we aim to select a subset of informative tokens to serve as input to the MLLM. The key insight is that adjacent frames in a video stream often contain redundant information, and selectively retaining only the most important tokens can significantly reduce the computational cost without compromising the reasoning accuracy.

We start by identifying a key frame that is most representative of the salient content in the temporal context. We compute the average saliency score for each frame and select the frame with the highest average score as the key frame:
\begin{align}
& \mathbf{P}_{t-M:t}^{\text{avg}}=\operatorname{Avg}(\mathbf{P}_{t-M:t})\in\mathbb{R}^{M+1} \\
& k = \operatorname{argmax}(\mathbf{P}_{t-M:t}^{\text{avg}})
\end{align}
where $\mathbf{P}_{t-M:t}^{\text{avg}}$ is a vector containing the average saliency scores for each frame in the visual context, and $k$ is the index of the key frame.

\subsubsection{Token Pruning}
Next, we compute a pruning score for each visual token in the non-key frames. The pruning score is a combination of two factors: (1) the saliency score of the token and (2) the dissimilarity between the token and its corresponding token in the key frame. The intuition is to retain tokens that are both salient and contain non-redundant information compared to the key frame.

Formally, let $\mathbf{P}_k\in\mathbb{R}^{N\times C}$ denote the feature map of the key frame and $\mathbf{P}_{\ne k}\in\mathbb{R}^{M\times N \times C}$ denote the feature maps of the non-key frames. We compute the cosine similarity between each token in the non-key frames and its corresponding token in the key frame:
\begin{equation}
\mathbf{S} = 1 - \operatorname{Sim}(\mathbf{P}_k, \mathbf{P}_{\ne k})\in\mathbb{R}^{M\times N},
\end{equation}
where $\operatorname{Sim}(\cdot, \cdot)$ denotes the cosine similarity function applied along the channel dimension.

The pruning score for each token is computed by adding the dissimilarity score $\mathbf{S}$ and the saliency score $\mathbf{P}_{\ne k}^{\text{avg}}$ obtained by averaging the feature map along the channel dimension:
\begin{align}
& \mathbf{P}_{\ne k}^{\text{avg}}=\operatorname{Avg}(\mathbf{P}_{\ne k})\in\mathbb{R}^{M\times N} \\
& \mathbf{I}_{\ne k} = \operatorname{argsort}(\operatorname{flatten}(\mathbf{S} + \mathbf{P}_{\ne k}^{\text{avg}}))
\end{align}

To perform token pruning, we sort the pruning scores in descending order and select the top $s$ percent tokens from the non-key frames, denoted as $\mathbf{I}_{\ne k}^{\text{keep}}\in\mathbb{R}^{s\times M\times N}$. The final set of selected tokens is obtained by concatenating the tokens from the key frame $\mathbf{I}_{k}$ and the selected tokens from the non-key frames $\mathbf{I}_{\ne k}^{\text{keep}}$.
The token pruning process reduces the total number of visual tokens from $(M+1)\times N$ to $(s\times M + 1)\times N$, resulting in a significant reduction in computational cost. By adaptively selecting the most informative tokens based on saliency and redundancy, our approach strikes a balance between efficiency and preservation of relevant information.

\subsection{Training and Inference}
We employ a visual instruction tuning approach~\cite{liu_visual_2023} to fine-tune the MLLM on domain-specific driving tasks while leveraging the knowledge from pre-trained vision and language models. The training data consists of video-question-answer triplets, where the answer includes both the textual response and, when applicable, the grounded object location represented by bounding boxes.

To encode the bounding box information, we follow the approach proposed in Qwen-VL~\cite{bai_qwen-vl_2023}. The bounding box coordinates are normalized to the range $[0, 1000]$ and formatted as a string ``[x1, y1, x2, y2]". Special ``box" tokens are added to the beginning and end of the string to distinguish it from regular text. Additionally, ``ref" tokens are used to associate the bounding box with the corresponding caption in the text.

During training, we first apply our proposed VTS algorithm to obtain the pruned visual tokens $\mathbf{H}'=\mathfrak{g}(\mathbf{X})[\operatorname{VTS}(\mathbf{X})]$, where $[\cdot]$ denotes the selection operation based on the token indices. The MLLM is then fine-tuned using the standard autoregressive language modeling objective, which maximizes the likelihood of the target response $\mathbf{Y}$ conditioned on the pruned visual tokens $\mathbf{H}'$ and the input question $\mathbf{Q}$:
\begin{equation}
\Pr(\mathbf{Y}|\mathbf{H}',\mathbf{Q})=\prod_{i=1}^{L}\Pr(y_i|\mathbf{H}',\mathbf{Q}_{<i},\mathbf{Y}_{<i};\theta)
\end{equation}

Here, $\theta$ denotes the trainable parameters of the LLM, $\mathbf{Q}_{<i}$ and $\mathbf{Y}_{<i}$ represent the tokens in the question and response preceding the $i$-th token, and $L$ is the total sequence length.

During inference, we apply the same VTS algorithm to prune the visual tokens and provide the pruned tokens along with the input question to the fine-tuned MLLM. The MLLM generates the textual response and the bounding box predictions autoregressively. 

Our VTS approach enables the efficient integration of temporal context into the reasoning process of MLLMs for autonomous driving tasks. By adaptively pruning redundant and less informative visual tokens, we significantly reduce the computational cost and memory requirements while preserving the most salient information. The fine-tuned MLLM can then leverage this compact yet informative representation to generate accurate and grounded responses, paving the way for more efficient language-based interaction and decision-making in autonomous driving systems.

\section{Experiments and Results}

\begin{table*}[!t]
    \centering
    \small
    \resizebox{1.0\linewidth}{!}{
    \begin{tabular}{l||cc|cccc|cc}
        \hline
        Method & \#Frames & \#Tokens & Lingo-Judge $\uparrow$ & BLEU-4 $\uparrow$ & METEOR $\uparrow$ & CIDEr $\uparrow$ & Memory (GB) & Throughput (RPM) \\
        \hline
        \hline
        \multirow{4}{10em}{LingoQA Baseline~\cite{marcu_lingoqa_2024}} 
        & 1 & - & 57.0 & 14.2 & 18.4 & 59.5 & - & -\\
        & 3 & - & 59.8 & 14.6 & 18.4 & 62.6 & - & -\\
        & 5 & - & 60.8 & 15.0 & 18.6 & 65.6 & - & -\\
        & 7 & - & 60.6 & 14.5 & 18.6 & 61.8 & - & -\\
        \hline
        GPT-4V Zero-Shot~\cite{marcu_lingoqa_2024,openai_gpt-4_2023} 
        & 5 & - & 59.6 & 6.3 & 12.4 & 42.8 & - & -\\
        \hline
        \hline
        Baseline~\cite{chen_internvl_2024} & 5 & 1280 & \textbf{64.2} & \textbf{15.6} & \textbf{20.5} & \textbf{57.8} & 58.4 & 126\\ 
        Temporal Res. ($0.6\times$)~\cite{chen_internvl_2024} & 3 & 768 & \underline{63.6} & 14.9 & \underline{20.4} & 54.4 & 42.0 & 180\\ 
        Spatial Res. ($0.75\times$)~\cite{chen_internvl_2024} & 5 & 720 & 61.4 & 13.9 & 19.1 & 51.0 & 40.6 & 174 \\ 
        ToMe~\cite{bolya_token_2023} & 5 & 720 & 62.0 & 14.2 & 19.7 & 52.1 & 40.6 & 180 \\ 
        PaPr~\cite{mahmud_papr_2024} & 5 & 768 & 62.8 & \underline{15.3} & 20.1 & 55.8& 42.1 & 174 \\ 
        VTS (Ours) & 5 & 768 & \textbf{64.2} & 14.5 & \textbf{20.5} & \underline{56.9} & 42.2 & 168 \\ 
        \hline
    \end{tabular}
    }
    \vspace{-1ex}
    \caption{\textbf{Performance Comparison on the LingoQA Dataset}. The best-performing method for each metric is highlighted in \textbf{bold}, while the second-best method is indicated by an \underline{underline}. $\downarrow$ denotes metrics where lower values are better, and $\uparrow$ indicates metrics where higher values are preferred. ``-" represents values that are not available for proprietary models. ``Res." stands for Resolution. Our proposed method, VTS (Video Token Sparsification), achieves competitive performance scores while significantly improving inference throughput and reducing the total number of visual tokens compared to the baseline method.} 
    \label{tab:sota-lingoqa}
\end{table*}

\begin{table*}[!t]
    \centering
    \small
    \resizebox{\linewidth}{!}{
    \begin{tabular}{l||cccccc|cc|cc}
        \hline
        Method & BLEU-1 $\uparrow$ & BLEU-4 $\uparrow$ & METEOR $\uparrow$ & ROGUE $\uparrow$ & CIDEr $\uparrow$ & SPICE $\uparrow$ & Mean-IoU $\uparrow$ & ACC $\uparrow$ & Memory (GB) & Throughput (RPM)\\
        \hline
        \hline
        LCP~\cite{malla_drama_2023} & 73.9 & \underline{54.7} & 39.1 & 70.0 & \textbf{3.7} & 56.0 & 61.4 & 68.4 & - & - \\
        \hline
        ToMe~\cite{bolya_token_2023} & \underline{74.2} & 53.6 & \underline{40.3} & \textbf{75.8} & \underline{2.9} & \textbf{58.2} & 59.2 & 66.7 & 42.3 & 186 \\
        PaPr~\cite{mahmud_papr_2024} & 74.1 & 53.1 & 39.7 & \underline{75.2} & 2.8 & 57.1 & \underline{65.7} & \underline{73.5} & 44.5 & 168\\
        VTS (Ours) & \textbf{75.3} & \textbf{55.8} & \textbf{40.7} & 74.7 & 2.8 & \underline{58.0} & \textbf{66.8} & \textbf{74.4} & 44.6 & 168\\
        \hline
    \end{tabular}
    }
    \vspace{-1ex}
    \caption{\textbf{Performance Comparison on the DRAMA Dataset.} 
    Our proposed VTS method achieves superior performance with competitive throughput and memory usage compared to other token reduction methods.}
    \label{tab:sota-drama}
    \vspace{-3mm}
\end{table*}

\subsection{Datasets}
We conduct comprehensive experiments on two autonomous driving video question-answering (VQA) benchmarks: DRAMA~\cite{malla_drama_2023} and LingoQA~\cite{marcu_lingoqa_2024}.
The DRAMA dataset consists of 17K scenario clips, each lasting 2 seconds, captured in highly interactive urban traffic scenes in Tokyo. The dataset includes various annotations, such as video-level QA, object-level QA, risk object bounding boxes, free-form captions, and separate labels for ego-car intention, scene classification, and suggestions to the driver. The 17K risk scenarios contain 12.3K vehicles, 3.3K pedestrians/cyclists, and 1.4K relevant traffic infrastructure elements. 

LingoQA is a benchmark dataset for video question answering in autonomous driving, collected in London, and designed to evaluate a wide range of skills. It consists of 28K unique scenarios, each lasting 4 seconds, and includes 419K annotations covering tasks such as description, counting, localization, anticipation, attention, and action justification. These datasets offer a comprehensive and challenging testbed for assessing the performance of our proposed approach in real-world autonomous driving scenarios.

\subsection{Evaluation Metrics}
We adopt several widely used language-based metrics for evaluating question-answering models in autonomous driving, following previous works~\cite{malla_drama_2023,marcu_lingoqa_2024}. These metrics include BLEU~\cite{papineni_bleu_2002}, METEOR~\cite{banerjee_meteor_2005}, ROUGE~\cite{lin_rouge_2004}, CIDEr~\cite{vedantam_cider_2015}, and SPICE~\cite{anderson_spice_2016}. While these metrics have known limitations, such as relying heavily on n-gram frequency rather than the underlying meaning of the answer, they provide a standardized way to compare the performance of different methods.

For the evaluation of object grounding on the DRAMA dataset, we use the Mean Intersection Over Union (Mean-IOU) and accuracy for $\text{IOU}>0.5$. These metrics assess the model's ability to accurately localize and associate the generated answers with the relevant objects in the input video frames.
On the LingoQA dataset, we use the official Lingo-Judge~\cite{marcu_lingoqa_2024} as the main metric for evaluation. Lingo-Judge is a small transformer-based text classifier that takes a question, the human's answer, and the model's answer as input and outputs a probability that the model's answer is correct. For every question, we run Lingo-Judge on all combinations of (human answer, predicted answer) and take the maximum correctness estimate. This metric provides a more human-aligned assessment of the model's ability to generate accurate answers.

\subsection{Implementation Details}
We adopt InternVL2-8B and InternVL2-2B~\cite{chen_internvl_2024} as the pretrained checkpoints for the LingoQA and DRAMA datasets, respectively. For the CNN proposal model $\mathfrak{p}$, we use MobileOne-S0~\cite{vasu_mobileone_2023}, which achieves a latency of 1ms on an iPhone 12, following the work of~\cite{mahmud_papr_2024}. To fine-tune the MLLM, we employ the LoRA (Low-Rank Adaptation) technique~\cite{hu_lora_2022}. LoRA learns a low-rank update to the pre-trained weights, significantly reducing the number of trainable parameters while achieving comparable performance to full fine-tuning. We set the rank $r$ to 512 and 256 for the LingoQA and DRAMA datasets, respectively, based on empirical observations. During the supervised fine-tuning stage, we keep the visual encoder, MLP projection layers, and the CNN proposal model fixed, focusing on adapting the LLM, which allows us to leverage the knowledge captured by the pre-trained models while efficiently specializing the MLLM for the autonomous driving domain. 

\subsection{VQA Results}
We compare the proposed VTS approach with other baseline approaches and state-of-the-art token reduction techniques. \cref{tab:sota-lingoqa} presents the comparison results on the LingoQA benchmark. The LingoQA Baseline is a closed-source model developed by Wayve~\cite{marcu_lingoqa_2024}. The baseline approach is the fine-tuned InternVL-8B model using all five frames provided. The Temporal Res. $(0.6\times)$ baseline uses three randomly selected frames out of the five. The Spatial Res. $(0.75\times)$ baseline reduces the spatial resolution by 1/4. ToMe~\cite{bolya_token_2023} and PaPr~\cite{mahmud_papr_2024} are two state-of-the-art token reduction approaches for Vision Transformers. The ``\#Tokens" column indicates the number of visual tokens input to the LLM. VTS uses 60\% of the visual tokens yet demonstrates competitive performance compared to the baseline, with a 28\% reduction in memory usage and a 33\% improvement in throughput.

\begin{figure}[!t]
\centering
\includegraphics[width=0.9\linewidth]{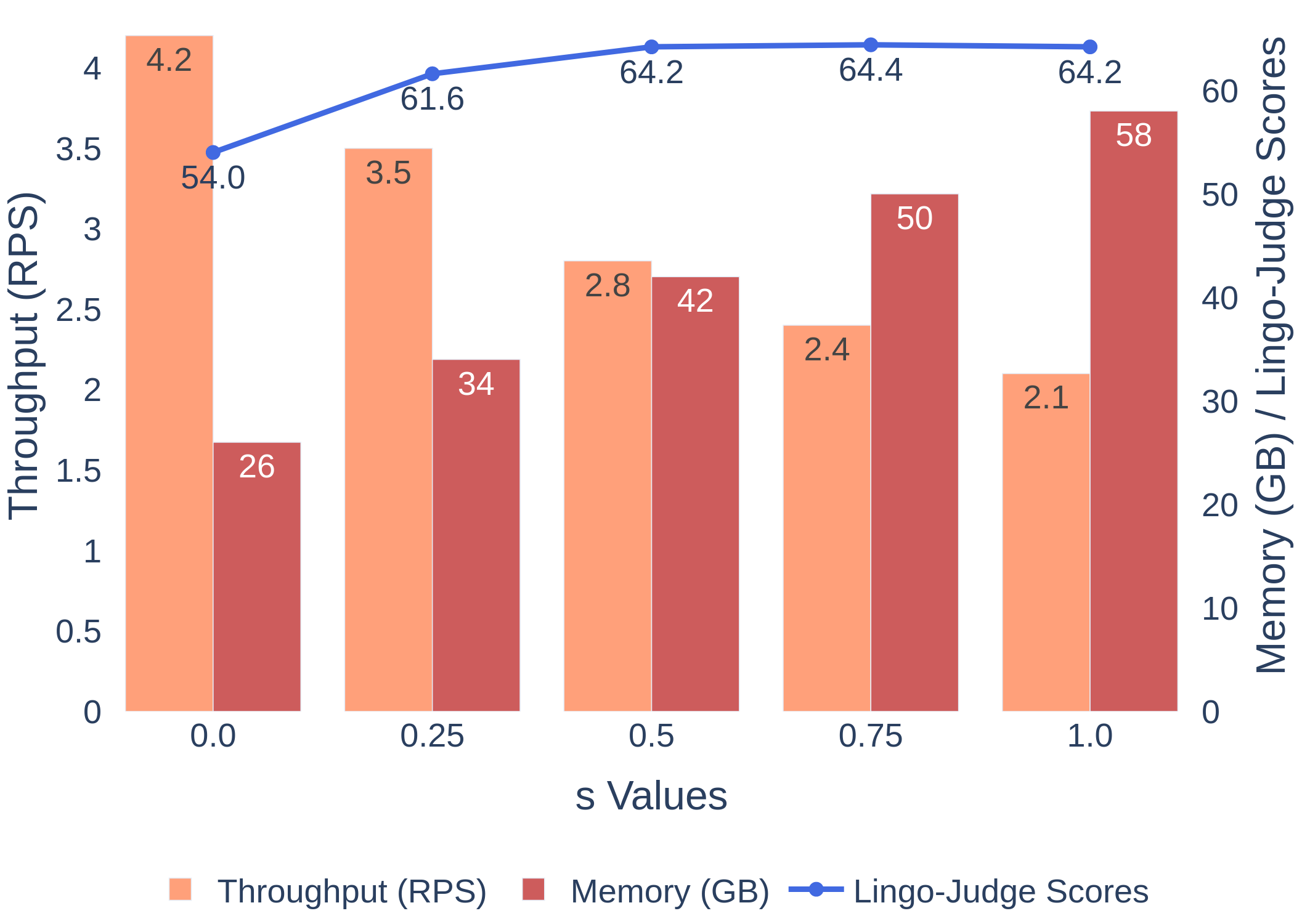}
\caption{Impact of token sparsification rate $s$ on inference throughput (request per second), memory consumption, and Lingo-Judge scores.}
\vspace{-3mm}
\label{fig:ab-s}
\end{figure}

\cref{tab:sota-drama} shows the comparison results on the DRAMA dataset. LCP~\cite{malla_drama_2023} is a state-of-the-art method specifically designed for the DRAMA dataset. VTS achieves state-of-the-art performance on the DRAMA dataset, outperforming LCP and other token reduction approaches on most metrics. Specifically, VTS obtains the highest scores on BLEU-1, BLEU-4, METEOR, Mean-IoU, and ACC. Notably, VTS maintains high performance while having memory usage and throughput comparable to PaPr, indicating its efficiency in processing video data. 

Overall, the main results showcase the superior performance and efficiency of VTS compared to state-of-the-art methods on two challenging autonomous driving video QA benchmarks. The proposed approach offers a promising direction for leveraging MLLMs in resource-constrained environments while maintaining high-quality visual reasoning capabilities.

\subsection{Ablation Study}
\begin{figure*}[!t]
\centering
\includegraphics[trim={5cm 0 0 0},clip,width=0.49\linewidth]{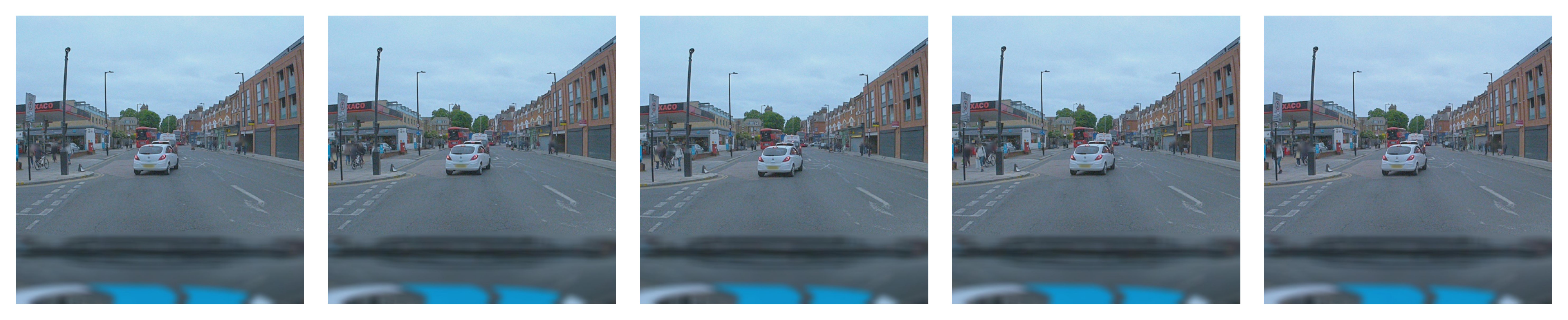}
\includegraphics[trim={0 0 5cm 0},clip,width=0.49\linewidth]{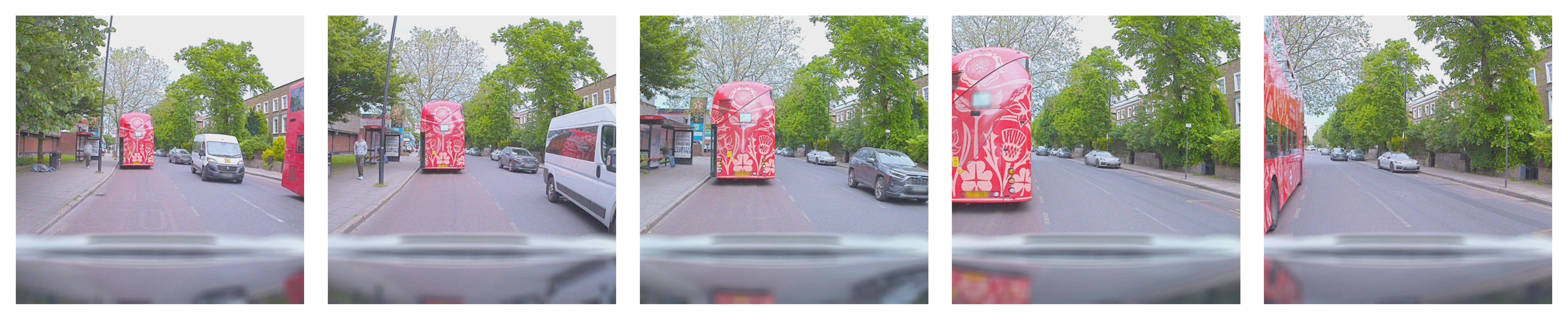}
(a) Images~\cite{marcu_lingoqa_2024}\\
\includegraphics[trim={5cm 0 0 0},clip,width=0.49\linewidth]{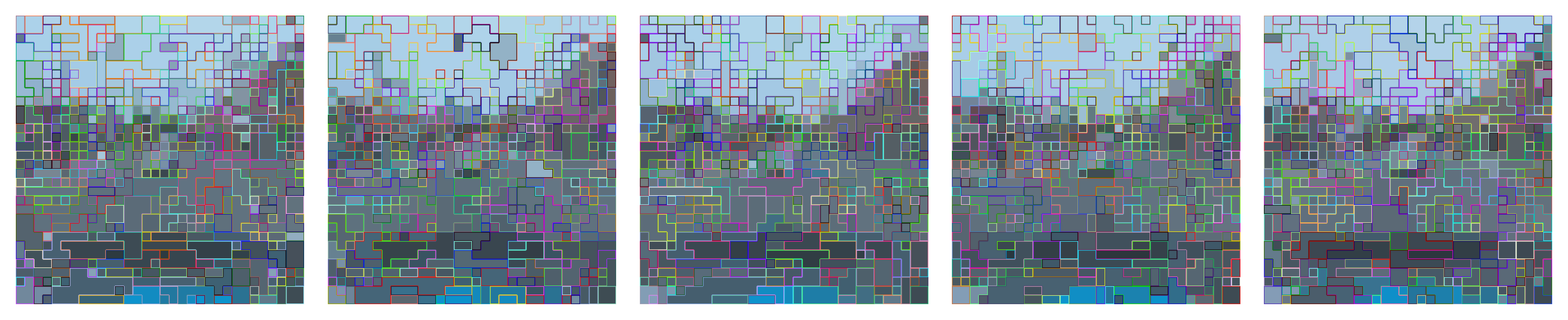}
\includegraphics[trim={0 0 5cm 0},clip,width=0.49\linewidth]{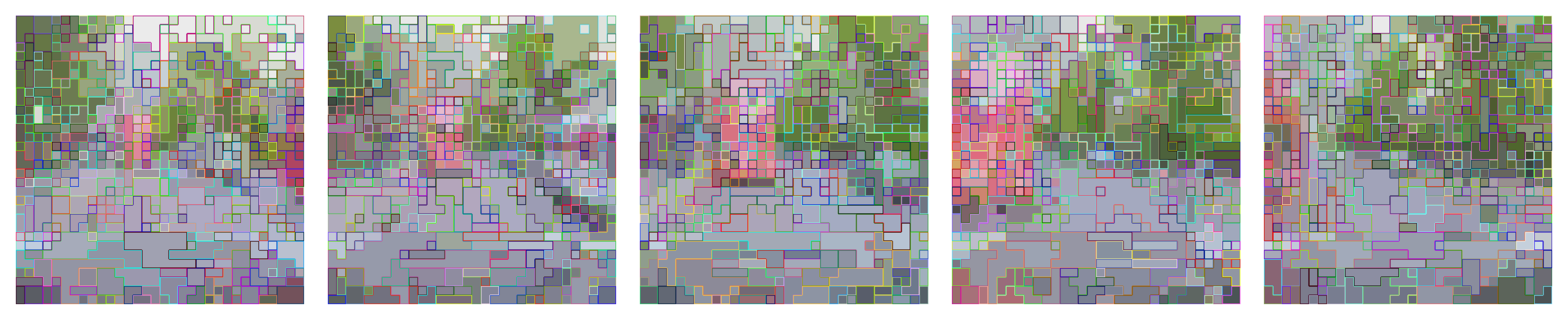}
(b) ToMe~\cite{bolya_token_2023}\\
\includegraphics[trim={5cm 0 0 0},clip,width=0.49\linewidth]{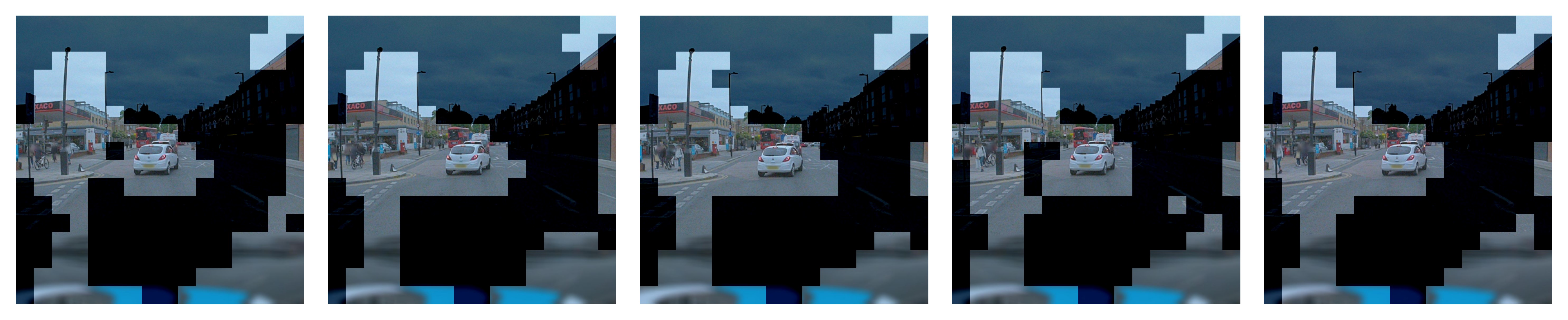}
\includegraphics[trim={0 0 5cm 0},clip,width=0.49\linewidth]{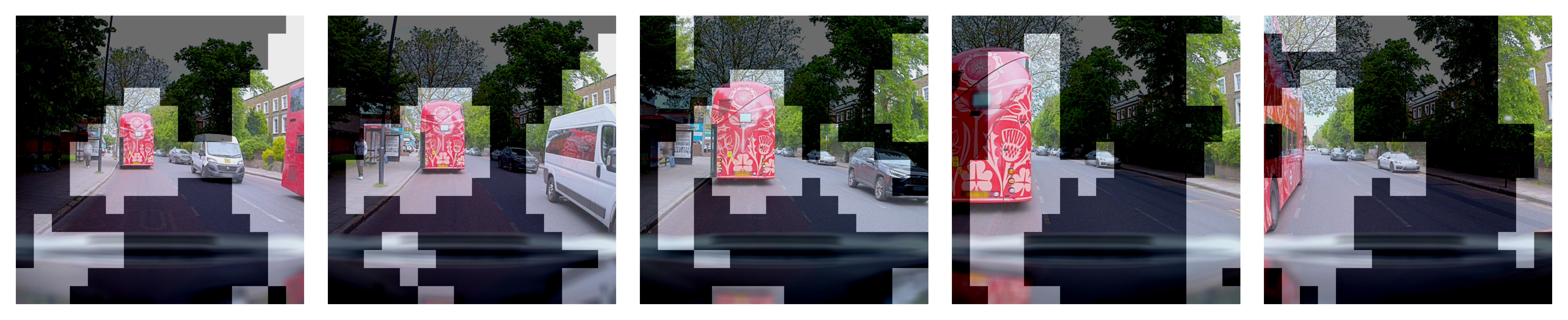}
(c) PaPr~\cite{mahmud_papr_2024}\\
\includegraphics[trim={5cm 0 0 0},clip,width=0.49\linewidth]{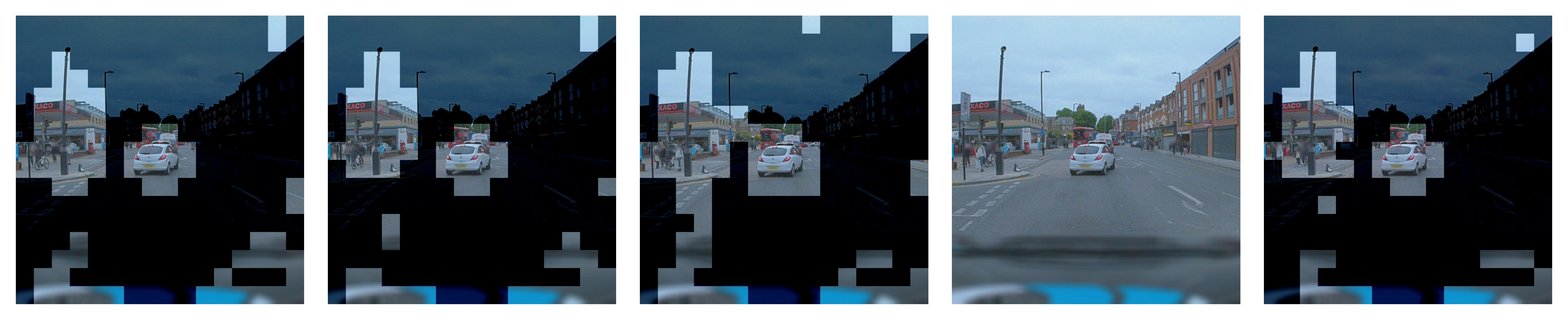}
\includegraphics[trim={0 0 5cm 0},clip,width=0.49\linewidth]{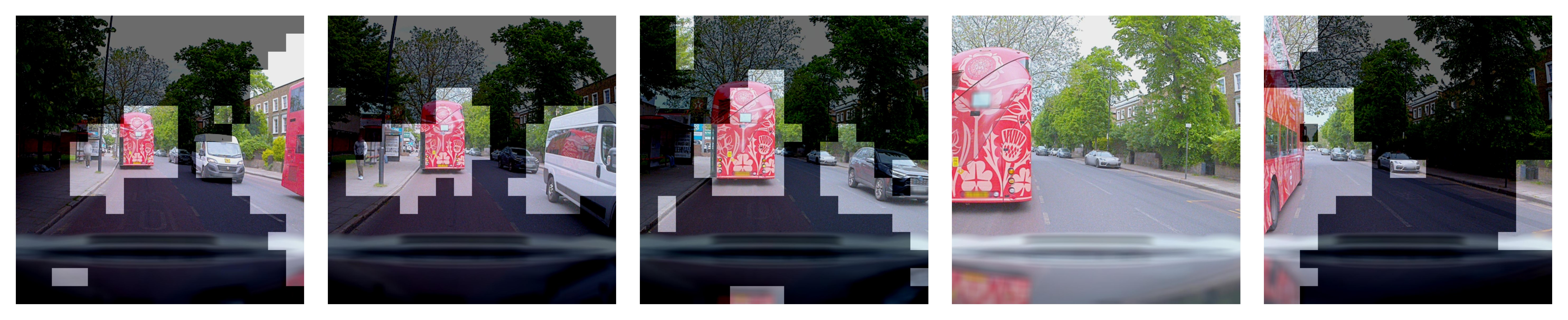}
(d) VTS (Ours)\\
\caption{Visualization of VTS compared with other token-reduction-based approaches.}
\vspace{-3mm}
\label{fig:vis-tr}
\end{figure*}

To investigate the impact of various components of our method, we conduct ablation studies on the LingoQA dataset. We examine the effectiveness of supervised fine-tuning, the key frame selection approach, and the impact of the token sparsification rate.
\paragraph{Effectiveness of Key Frame Selection}
We compare our adaptive key frame selection algorithm with a random key frame selection baseline. Given a sequence of consecutive images, the baseline randomly selects a key frame and then applies the proposed VTS algorithm. As shown in \cref{tab:ab}, the adaptive key frame selection method consistently outperforms the random key frame baseline, both with and without fine-tuning, in terms of the Lingo-Judge score. This highlights the importance of selecting the most informative frame as the key frame for effective token sparsification.

\begin{table}[!t]
    \centering
    \small
    \resizebox{\linewidth}{!}{
    \begin{tabular}{ccc|cccc}
        \hline
        Key Frame & \#Tokens & FT & LJ $\uparrow$ & BLEU-4 $\uparrow$ & METEOR $\uparrow$ & CIDEr $\uparrow$\\
        \hline
        \hline
        All & 1280 & \xmark & \textbf{55.8} & \textbf{5.7} & \textbf{10.4} & \textbf{14.3}\\
        Random & 768 & \xmark & 51.8 & \underline{5.0} & \underline{9.7} & 13.5\\
        Adaptive & 768 & \xmark & \underline{52.2} & 4.8 & 9.5 & \underline{14.0}\\
        \hline
        All & 1280 & \cmark & \textbf{64.2} & \textbf{15.6} & \textbf{20.5} & \textbf{57.8} \\ 
        Random & 768 & \cmark & \underline{63.4} & \underline{14.7} & \textbf{20.5} & 53.2\\
        Adaptive & 768 & \cmark & \textbf{64.2} & 14.5 & \textbf{20.5} & \underline{56.9} \\
        \hline
    \end{tabular}
    }
    \vspace{-1ex}
    \caption{Ablation study comparing the performance of different key frame selection strategies with and without fine-tuning (FT) on the LingoQA dataset. The ``All" row represents using all visual tokens without any sparsification. ``Random" and ``Adaptive" refer to random key frame selection and our proposed adaptive key frame selection, respectively. The number of tokens used in each setting is also reported.}
    \vspace{-3mm}
    \label{tab:ab}
\end{table}

\paragraph{Impact of Supervised Fine-Tuning}
Our proposed approach supports training-free application, although it may suffer from some performance loss. For example, with a sparsification rate $s=0.5$, the Lingo-Judge score drops from 55.8 to 52.2 in the zero-shot setting (\cref{tab:ab}). This performance gap can be attributed to the altered distribution of visual tokens conditioned on by the MLLM after applying the VTS algorithm. However, supervised fine-tuning effectively closes this performance gap. After fine-tuning, the baseline method using all visual tokens achieves a Lingo-Judge score of 64.2, while our VTS approach reaches the same score of 64.2. This indicates that VTS with fine-tuning can significantly reduce computation and memory requirements without sacrificing performance, validating the effectiveness of the VTS algorithm in retaining the most informative tokens and removing redundant ones.

\paragraph{Impact of Token Pruning Rate}
The token sparsification rate $s$ is a key hyperparameter in our VTS algorithm, significantly influencing inference throughput, memory consumption, and accuracy (\cref{fig:ab-s}). Memory consumption is linearly related to the $s$ value, ranging from 26GB at $s=0$ to 58GB at $s=1.0$. Inference throughput exhibits a roughly quadratic improvement with respect to $s$. Lower $s$ values (i.e., keeping fewer tokens) result in better throughput improvement. However, inference accuracy can also decrease with higher $s$ values. For $s>0.5$, the accuracy remains stable at around 64.3, but for $s<0.5$, performance starts to drop, indicating that keeping too few tokens inevitably leads to the loss of important information.

These ablation studies highlight the effectiveness of our key frame selection algorithm, the benefits of supervised fine-tuning, and the trade-off between computation efficiency and performance when adjusting the token sparsification rate. The insights gained from these experiments guide the selection of optimal hyperparameters and validate the design choices in our main results.
\begin{table*}[!t]\centering
\begin{minipage}{1.0\linewidth}\vspace{0mm}    
\centering
\begin{tcolorbox} 
\centering
\footnotesize
\begin{tabular}{p{0.99\columnwidth} c}
\includegraphics[width=0.85\textwidth]{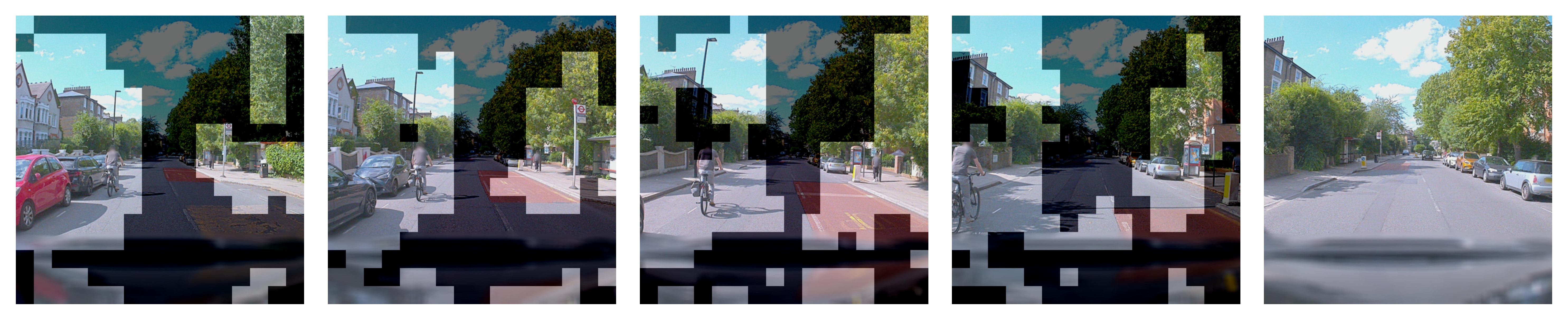} \\
\textbf{Query:} Can you identify the \hl{colors of the cars parked on the left side of the road}?
\textbf{Response (Human):} Yes, the identifiable cars parked on the left side of the road are \hl{a red sedan and two black sedans}. {\color{red}\textbf{Response (Baseline):}} \hl{The image does not provide enough detail} to confidently identify the colors of the cars parked on the left side of the road. {\color{blue}\textbf{Response (VTS):}} Based on the image, there are two cars parked on the left side of the road. \hl{The car closest to the camera is red, and the one further away is black}.

\hrulefill & \\
\includegraphics[width=0.85\textwidth]{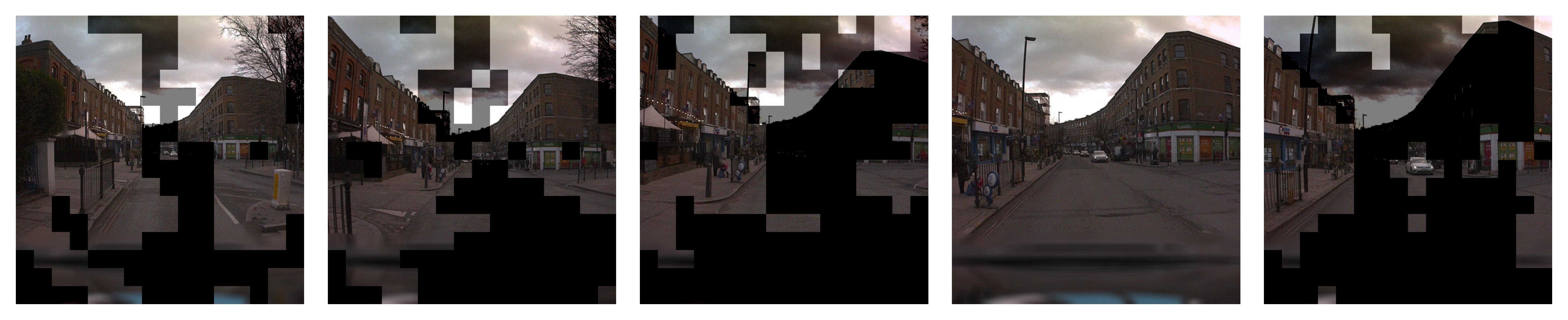} \\
\textbf{Query:} What is your current action and why? \textbf{Response (Ground truth):} I am following the lane and maintaining right to be correctly centered in the lane and to keep a safe distance from the kerb on the left. \hl{This is to ensure safe driving and avoid any potential collisions with other vehicles or objects on the road}. {\color{red}\textbf{Response (Baseline):}} As an autonomous vehicle driving in London, my current action is to navigate through the city streets, specifically the one depicted in the image. \hl{The reason for this action is to ensure safe and efficient transportation for passengers, while adhering to traffic rules and regulations. Here's a detailed breakdown of the actions and considerations: 1. **Navigation and Positioning**: I am currently positioned at the intersection of the street, preparing to move forward. The image shows a relatively quiet street with minimal traffic, which is ideal for smooth navigation. \textbf{[omitted]} 7. **Communication**: I can communicate with other vehicles and infrastructure, such as traffic lights and road signs, to optimize my route and ensure safe navigation. The image shows a clear road ahead, but I will remain prepared to respond to any signals or instructions.} In summary, my current action is to navigate through the city street, ensuring safety, adhering to traffic rules, and providing a comfortable ride for passengers. The image depicts a relatively quiet street, which allows for smooth and efficient movement. {\color{blue}\textbf{Response (VTS):}}
As an autonomous vehicle, my current action is to navigate through the urban environment of London. \hl{My primary goal is to reach my destination while adhering to traffic laws and avoiding any potential hazards}.

\end{tabular}
\end{tcolorbox}
\vspace{-2mm}
\end{minipage}
\caption{Zero-shot text generation with and without VTS.}
\vspace{-3mm}
\label{tab:h_example}
\end{table*}

\subsection{Visual Comparisons with other Methods}
We present a visual comparison of the proposed VTS with other token-reduction-based approaches in \cref{fig:vis-tr}. The token reduction rate is set to approximately 40\% for all methods to ensure a fair comparison. We showcase two randomly selected cases from the LingoQA dataset, each consisting of four consecutive frames sampled at a frequency of 1 Hz.

The first row of \cref{fig:vis-tr} shows the original image frames. The second row illustrates the results of the ToMe~\cite{bolya_token_2023} method, where patches sharing the same inner and border color are merged. The third and fourth rows depict the visualizations of PaPr~\cite{mahmud_papr_2024} and our proposed VTS approach, respectively. In these rows, black masks are applied to patches pruned by the corresponding algorithms, emphasizing the spatial distribution of the retained tokens. Upon closer inspection, VTS demonstrates its effectiveness by exploiting the inherent high sparsity of video frames, leading to a more informative and compact pruning compared to other techniques. The strength of VTS lies in its ability to select a key frame that establishes global context while identifying the most informative and dynamic patches across the remaining frames.

\subsection{Qualitative Analysis on Zero-Shot Inference}

We present a qualitative analysis of zero-shot text generation with and without VTS (referred to as Baseline) in \cref{tab:h_example}. The shown images are results with masked regions pruned by VTS, but for the Baseline, the original unpruned visual tokens are input to the MLLM for inference. The experiments are conducted on the LingoQA dataset using InternVL-8B without fine-tuning. From the results, we can draw several insights: (1) The generation is largely conditioned on the visual tokens, i.e., the visual tokens have a significant impact on the generation results. (2) The proposed VTS can work effectively in a zero-shot setting without fine-tuning and generate accurate reasoning with the pretrained MLLM. However, fine-tuning the MLLM with VTS further improves the performance, as demonstrated in our VQA results. (3) By reducing the number of visual tokens input to the LLM, VTS not only reduces the computational and memory cost but also has a positive impact on the generation quality, especially for video input. By pruning redundant and non-relevant visual tokens, VTS allows the generation to focus on the patches of interest, leading to more accurate and coherent responses.

In the first example, due to the small portion of tokens that are actually relevant to the text query (only the left part of the first two frames are relevant, while the other patches are noise), the Baseline model incorrectly determines that ``The image does not provide enough detail". However, with VTS, the model correctly identifies the colors of the cars by focusing on the relevant visual tokens. In the second example, where the query is more vague, the Baseline model hallucinates a lot of irrelevant information, while the VTS model generates a response more aligned with the human-expected answer. Overall, the qualitative analysis highlights the potential of VTS as a powerful technique for enhancing the performance of MLLMs in zero-shot inference.

\section{Conclusion}
In this paper, we introduced Video Token Sparsification (VTS), a novel approach to address the computational challenges of integrating multimodal large language models (MLLMs) into autonomous driving systems. By adaptively pruning less informative visual tokens while preserving the most salient information, VTS significantly reduces computational overhead and improves inference throughput without compromising performance on various video question answering tasks. Our experiments on the DRAMA and LingoQA benchmarks, along with qualitative analysis and visualizations, demonstrate the effectiveness of VTS in enabling efficient multimodal reasoning for resource-constrained onboard computing platforms. We believe that our work will inspire further research on efficient multimodal reasoning techniques and contribute to the development of more intelligent, explainable, and user-friendly autonomous driving systems.






\clearpage
{\small
\bibliographystyle{ieee_fullname}
\bibliography{ma,ref,egbib}
}

\end{document}